# From "Made In" to Mukokuseki: Exploring the Visual Perception of National Identity in Robots

**KATIE SEABORN**

*Industrial Engineering and Economics (IEE)*
*Tokyo Institute of Technology*

**HARUKI KOTANI**

*Industrial Engineering and Economics (IEE)*
*Tokyo Institute of Technology*

**PETER PENNEFATHER**

*gDial, Inc.*
*University of Toronto*

**ABSTRACT:**    People read human characteristics into the design of social robots, a visual process with socio-cultural implications. One factor may be nationality, a complex social characteristic that is linked to ethnicity, culture, and other factors of identity that can be embedded in the visual design of robots. Guided by social identity theory (SIT), we explored the notion of "mukokuseki," a visual design characteristic defined by the absence of visual cues to national and ethnic identity in Japanese cultural exports. In a two-phase categorization study (n=212), American (n=110) and Japanese (n=92) participants rated a random selection of nine robot stimuli from America and Japan, plus multinational Pepper. We found evidence of made-in and two kinds of mukokuseki effects. We offer suggestions for the visual design of mukokuseki robots that may interact with people from diverse backgrounds. Our findings have implications for robots and social identity, the viability of robotic exports, and the use of robots internationally.

**DEFINITIONS:**

*Mukokuseki:*    Literally "stateless," "no nationality," or "nationless," here refers to no perceivable cue to national origin or identity.

*Takokuseki:*    Literally "multiple nationalities," here refers to plurality in perceptions of national origin or identity.

*Made-in effect:*    National origin or identity is identifiable.

# 1   Introduction

Robots are being taken up in a variety of contexts with a diverse array of people around the world. Robots produced in one country are being exported to other countries at small and large scales. The collaboration between Aldebaran in France and SoftBank in Japan, for instance, led to the development of Pepper, one of the most recognizable social robots used in research and daily life. Many robots, like Pepper, have a humanoid form or feature some level of anthropomorphism in their visual design. These visual cues may be "designed in" to the robot's appearance on purpose *or* without the intention (or realization) of the designer. Still, they are perceivable and interpreted by people in relation to human models and especially stereotypes, often unconsciously but with implications for the interaction with the robot [22, 39, 43]. This highlights a modern challenge in reconciling the increasing "entanglement" of social and material worlds [21], where we and possibly our robotic creations are participating in social categorization, identification, and comparison activities, i.e., social identity dynamics [27, 49]. Understanding whether and how these visual cues and resulting categorizations influence people's attitudes and behaviour towards robots has become a key area of interest within industry and the academy, especially in light of recent calls for action on designing for diversity within robotics and adjacent spaces [23, 47, 53, 62].

At a fundamental level, the visual design of robots, especially socially interactive, anthropomorphic ones, may be understood through the socio-cultural context in which the robot was designed and wherein the person is interacting with the given robot. Visual cues to humanlikeness that are embedded, purposefully or not, in the physical design of a robot may be read as human social characteristics [14, 54]. These may then be linked to culturally-mediated models of human social groups, a phenomenon captured under the concept of social identity theory (SIT) [27, 60, 61, 64]. Such social categories and characteristics include gender [9, 19, 41, 50], race and ethnicity [7, 35, 52], nationality [19, 33, 54], emotional expression [20, 28, 58], nonverbal behaviours [13, 45, 65], and more. The degree to which people key into these visual cues as social categories and under what circumstances with robots is an ongoing question. Relatedly, the degree to which designers and roboticists consciously embed these visual cues, as well as to what extent researchers, educators, and practitioners make conscious choices about employing robots based on these visual cues, is typically unknown, with some evidence to suggest this is unconsciously done or without awareness of the effects. A critical review of Pepper, for instance, found a confusion of gender attributions by researchers and participants based on research reporting [50].

Another possibility is that these cues are vague by design. *Mukokuseki* is a Japanese concept about the visual design of characters and commodities whereby no trace of nationality, race, ethnicity, or culture is (meant to be) perceivable: effectively, the appearance is "stateless," a socio-cultural tabula rasa [1, 31, 32, 44]. A mukokuseki approach may be purposefully employed, simply assumed, or an unconscious result of the design process. Indeed, many products and cultural exports from Japan rely on the assumption of mukokuseki for economic success internationally. Yet, it is not clear whether and to what degree this has been done or holds true for robots, especially in the presence of intersecting factors, such as anthropomorphism. Moreover, mukokuseki has implications for diversity and representation, given that many robots are designed with forms and shades that may be perceived as racial or ethnic cues [7, 53]. Even if designers purposefully employ a mukokuseki approach or assume that a given robot is mukokuseki by design, this is not necessarily the case for end-users. When it comes to user experience (UX), experts must ultimately accept that end-users may interpret their creations differently than intended or expected [16]. The question of whether robots are truly mukokuseki or possibly betray their "made in" origins [54] through visual cues un/knowingly embedded in their morphologies is an open question.

To this end, we conducted a categorization study to evaluate whether and how robots were deemed mukokuseki based solely on their visual design, i.e., using photos of robots as visual stimuli. We asked: ***Does the phenomenon of mukokuseki occur for robots?*** We recruited two cohorts from nations associated with robots in industry and daily life—America and Japan—to rate a series of nine robots from each nation. Drawing from SIT, we asked: ***RQ1: Do people interpret a robot's appearance in line with its origins, i.e., are there perceivable "made in" visual cues, or do they read in their own or no national, ethnic, and/or cultural background, i.e., a mukokuseki effect?*** Nationality is a human concept and may be invoked by relatively more humanlike designs. As such, we also varied the selection of robots by level of anthropomorphism, asking: ***RQ2: Does the degree of anthropomorphism influence the presence or absence of a mukokuseki effect?*** Our main contributions are empirical, cross-cultural evidence of both mukokuseki and made-in effects that were not universal across the robots, national cohorts, or individuals, as well as the presence of two forms of mukokuseki. A high degree of perceived anthropomorphism and individual perceptions of similarity appear to play key roles. Our findings have implications for robotics research and practice, especially for visual design and cross-cultural initiatives.

## 2  Background

### 2.1  MUKOKUSEKI AND NATIONALITY, RACE, ETHNICITY, AND CULTURE

People are diverse, and arguably our robotic creations should reflect that plurality. The challenge of designing social robots for diversity and a multiplicity of perspectives is grounded in the need to anticipate people's propensity to anthropomorphize machines and other objects. When people interact with non-human agents, such as robots, that are designed to behave as if they can act with intelligence, foresight and intentions, they will anthropomorphize the robot [6, 17, 39, 43]. We look for and interpret humanlike cues—not only visual, but also auditory, gestural, and other modes of embodiment—as a means of understanding what the agent is and how it may behave. There can be individual variability as well as group variability at all scales, and these may intersect in complex and unexpected ways. Nevertheless, anthropomorphism is generally considered an accessible strategy for understanding and anticipating socially intelligent robotic behaviour.

One aspect of anthropomorphism long overlooked relates to the robot's "origins." We capture this under the heading of "nationality," a complex characteristic of people that refers to one's legal and psycho-physical location/s within the human-occupied world of nations and boundaries, as well as cultural background, of which there may be multiple, and racial and/or ethnic diversity in appearance, behaviour, attitudes, social expression, social status, and more [3, 10, 12]. The degree to which nationality is tied to and intersects with culture, race, and ethnicity can vary widely within and across nations as well as individual people. Notably, these are all social constructs [52, 63], categories we have created to organize ourselves. For instance, critical historians have argued that "race" was constructed to justify certain acts on and the removal of liberties for groups of people who shared certain physical characteristics, cultures, and/or national origins [24]. More subtly, we may assume that the more socially powerful groups represent everyone, i.e., white American men are the "default" [53, 56, 59], or attempt to evade the issue, despite un/known effects [5]. Growing recognition of social power and representation as a matter of fairness has led to calls for critical reflection and engagement [47, 48, 53].

Mukokuseki offers another perspective. Iwabuchi [31] coined the term to describe the phenomena of the transnational success of Japanese electronic consumer goods based on national "vagueness" in the design of the visual appearance of the characters. "Mu" means absence, while "kokuseki" means nationality, i.e., the absence of nationality. In short, mukokuseki is the notion that the visual designs of such cultural outputs have no cues to nationality. This may be *by design*, i.e., a conscious design choice; *by assumption*, i.e., a premise of the design; and/or *a perception*, i.e., a consequence of how the design is interpreted. According to Iwabuchi, mukokuseki has two explicit premises: *statelessness*, or lacking cues to national identity, and *trans-cultural adaptivity*, whereby anyone from any nation or culture can readily accept the goods as their own. This potential for a plurality, rather than an absence, of national identity prescriptions was raised by Adamowicz [1], who suggested a complementary term: *takokuseki*, where instead of "mu" we have "ta" to mean "multiplicity." As implied by Iwabuchi, we cannot know whether mukokuseki is a conscious design choice or assumption without asking the designers, and we typically do not have this information. Designers may also not employ mukokuseki consciously. In any case, whether mukokuseki has been achieved—no cues to nationality plus adoptability by anyone from anywhere—is ultimately the domain of end-user perceptions [16]. This is what motivated us to start our exploration with a user perceptions study, rather than interviews with designers.

We must also consider a hidden premise about what "nationality" can mean: not only national identity, but also cultural, racial, and/or ethnic identity. Japan, for instance, is not a mononational, monocultural, monoracial, or monoethnic nation. Ethnic groups with Japanese nationality include the Burakku, Indigenous Ainu, zainichi Koreans, as well as immigrants from around the world. Yet, the national identity of Japan is largely constructed in a "mono" way around the majority ethnic population: the Yamato or Wajin [38]. This, alongside the current global context of who has social power, notably histories of white Western imperialism and colonialism, and the post-war influence of the West, especially the US, on Japan [46], raises questions about whether and for whom mukokuseki occurs. Notably, while mukokuseki emerged from Japanese theorizing about Japanese cultural exports, it may be a general concept that has potential applications elsewhere, i.e., for robots and other nations.

Indeed, a small but growing body of work has discovered implications of nationality, culture, and race and/or ethnicity for robots. Spatola et al. [54] primed people to key into human stereotypes about national identities by labeling certain robots as having certain national identities. These stereotypes, such as about national warmth, held true for robots, indicating that a "made in" effect could be induced simply by assigning national identities to any robot. We go one step further by refraining from priming the national identities explicitly, exploring whether robots, in the absence of primes, are "stateless" or mukokuseki. Marchesei, Roselli, and Wykowska [36] considered how the national identity and cultural orientations towards collectivism influenced perceptions of a robot, finding that individual differences favouring collectivism influenced acceptance of the robot as a member of the social group. We add on by considering individual and cultural-level possibilities. Bartneck et al. [7] replicated the study by Eberhardt et al. [18] on race association with gun violence, which involved participants indicating the moment that they could identify the content of an image—a gun—that was slowly changing from blurry to clear. In the robot study, robots with darker "skins" experiencing a similar "shooter bias" as in the original study, where white participants more quickly identified the image as a gun after being primed by a split-second image of a Black face, compared to a white face or an abstract shape. This suggests that visual design choices, such as apparent race, can have implications that match human models, further raising doubts about the "mukokuseki-ness" of robots.

## 2.2 MUKOKUSEKI AND SOCIAL IDENTITY THEORY

Nationality, race, ethnicity, and culture are social characteristics as well as matters of social identity. We can refer to the umbrella model of social identity theory (SIT) to help explain and predict responses to robots that may have these characteristics. SIT is based on the idea of the social categorization of self and others, a dynamic and process that is both personal and social. People tend to assign themselves to social groups (in-groups) and not others (out-groups). SIT posits that such sociological characteristics as national identity will influence a person's affinity and willingness to interact with others [27, 60, 61, 64].

Given that national identity is a human phenomenon, the degree of humanlikeness of the robot, i.e., its level of anthropomorphism, may impact the degree to which people pick up on and accept these social designations. As such, we needed to select our visual robotic stimuli by degree of anthropomorphism. Following Spatola et al. [54], we selected robots that, based on their appearance, would fall into one of four distinct categories of anthropomorphism offered by Duffy [17]: industrial, mechanical humanoid, iconic humanoid, and humanoid. We then conducted a manipulation check of our selection to ensure that each robot would in fact be generally prescribed by others as falling into each of these categories in the way that we expected (refer to 3.3.3).

We do not know whether and to what degree a mukokuseki effect exists for robots. The mukokuseki premise is that robots will not have a perceivable national identity. Still, given that it is a human characteristic, we may expect that robots with higher levels of perceived anthropomorphism, i.e., humanlike robots, would be viewed as having some form of human nationality as a premise of "being human." We thus hypothesized:

> **H1.** *Robots with a high degree of anthropomorphism, i.e., humanoid robots, will be perceived as having a human nationality, i.e., a made-in effect by anthropomorphism.*

The other work reviewed, notably the report by the non-Japanese developers of the iconic humanoid "designed for Japan" robot Pepper [40], suggests that any robot, as a designed object, may have intentional or unintentional cues to national identity that could be perceived by people through its visual design, i.e., a made-in effect. We thus hypothesized:

> **H2.** *Robots made in a certain nation will be more often assigned that nationality, i.e., a made-in effect by design.*

Robots occupy a liminal space, where contradictions may be possible. Robots of any level of anthropomorphism or with any visual cue to national identity may simply be perceived as a "made in" object: produced within a nation but not deemed to be "of" that nation like people are. We can explore this possibility through similarity-to-self. SIT predicts that our perceived similarity to others is premised in social characteristics like national identity. Thus, a robot deemed by a person to be similar to them may also be deemed by that person as having the same national identity, i.e., being part of the same social in-group. We thus hypothesized:

> **H3.** *The perceived national identity of the robot will reflect that of the attributor when the attributor also perceives that robot as similar to themselves, i.e., a made-in effect by similarity-to-self.*

# 3    Methods

We conducted an online, cross-cultural, within-subjects categorization study, following similar research survey designs, e.g., [51, 54]. The within-subjects variable was the robot stimuli; all participants categorized all robots, which were presented in a random order. The cross-cultural aspect refers to the participant pools (American and Japan-based respondents) and the national origin of the robot stimuli (robots built in America and Japan, plus the Japan-France case of Pepper).

Two phases were conducted. In the first phase, the US iconic humanoid robot (Bandit) was deemed less anthropomorphic than expected and replaced by another robot (Octavia) in the second phase (refer to 3.3.3). Since both robots received similar ratings in the end, we ultimately combined the data sets from both phases, allowing us to double the number of samples. The basic protocol was registered at OSF before data collection on Oct. 6[th], 2021[1].

## 3.1    PARTICIPANTS

A total of 202 participants were involved over two phases (n=109, n=93; full details are in Table 1). Different participants pools were used for each phase and national group. In the first phase, 54 Americans were recruited through Amazon Mechanical Turk (AMT or mTurk) and 55 Japanese respondents were recruited through word-of-mouth sampling (n=109). In the second phase, 56 American and 37 Japanese respondents were recruited from Prolific. To ensure response quality, we used the random code procedure by Nicoletti[2], which required participants to complete the study and input a random code generated on the final page into the recruitment system to receive credit. mTurk participants needed the Masters Qualification, which is granted by Amazon when a worker has achieved high performance across a variety of tasks[3], and a HIT approval rate of over 95%. With checks in place, mTurk and Prolific are expected to produce responses comparable to within 10% of traditional survey panel methods [8]. Some respondents disclosed other national, ethnic, and cultural identities; we accepted this as a characteristic of identity plurality, grouping these participants with the rest of their cohort. Participants were compensated roughly $4 USD or ￥300. Ethics approval was granted by the institutional ethics board at Tokyo Institute of Technology (approval #2023358).

---

1  https://osf.io/ack48
2  http://nicholasnicoletti.com/survey-monkey-and-mechanical-turk-the-verification-code
3  https://www.mturk.com/worker/help

## 3.2  PROCEDURE

Participants were given a link to the English or Japanese version of the online questionnaire. The first page provided study details and asked for consent. Participants were then presented with an overview of all robot images in a random order. This was to create a baseline for relative comparisons even while evaluating robots individually. Participants were then presented with each individual robot in a random order. The instruments (refer to 3.4) were folded into a larger study alongside items about gender and age as distractors [15]. Next, participants were asked to write down the national origin of each robot, if they knew it. They then provided demographics and collected the code for completion and compensation. The study took an average of 20 minutes.

*Table 1: Participant demographics.*

| Phase | Nation | Gender | Age | Nationality |
|---|---|---|---|---|
| 1 | US | 33 men, 21 women, and no others reported | 11 aged 25-34, 21 aged 35-44, 11 aged 45-54, 6 aged 55-64, 4 aged 65-74, and one aged 75 or older | 54 Americans |
| 1 | JP | 29 men, 26 women, and no others reported | 53 in their 20s, 2 in their 50s | 55 Japanese |
| 2 | US | 28 men, 27 women (one trans*), and no others reported | 12 aged 18-14, 20 aged 25-34, 10 aged 35-44, 5 aged 45-54, 7 aged 55-64, and one aged 65-74 | 55 Americans |
| 2 | JP | 10 men, 27 women, and no others reported | 7 in their 20s, 13 in their 30s, 9 in their 40s, 6 in their 50s, and 2 in their 60s or above | 35 Japanese, 5 having dual citizenship (4 American, one N/A), and one other |

*Demographic totals may not match overall totals due to unreported demographics. Age formats differ due to SurveyMonkey defaults for each language.*

### 3.3   MATERIALS AND STIMULI

#### 3.3.1   *Questionnaires.*

We used the multilingual SurveyMonkey online platform to deploy the questionnaires. We used the original instruments or official translations where possible. Otherwise, a native Japanese speaker with advanced English ability and an advanced Japanese speaker native in English translated and back-translated materials with additional checks using Deepl[4], an imperfect but slightly higher quality tool than alternatives [26].

#### 3.3.2   *Stimuli Selection.*

Nine images of robots were used as visual stimuli[5]. Following previous research [51, 54], we used the Duffy designations [17] to select the robots, aiming for those formally recognized in the IEEE Robots database[6], and then conducted a manipulation check (refer to 3.3.3) to confirm our selections. We aimed to select similar-looking pairs of robots, one pair for each Duffy category, from the US and Japan. We were unable to do this for the mechanical and mechanical humanoid robots, so we searched Japanese robotics company websites for alternatives, finally selecting duAro and T-HR3. Two researchers finalized the selection after several rounds of discussion.

Five robots were made in America: Cody[7], CHARLI[8], Bandit[9], Octavia[10], and Diego-san[11]. Three robots were made in Japan: duAro[12], T-HR3[13], HRP-4C[14]. We also included the special case of Pepper[15], an internationally known and nationally ambiguous robot created by a French company (Aldebaran Robotics) in collaboration with a Japanese company (SoftBank Robotics) for Japanese and international markets. Pepper is often associated with Japan by way of this collaboration and SoftBank's local marketing efforts, leading to high cultural integration. Examples include the *Pepper Parlor* restaurant in Shibuya, Tokyo[16] and Pepper's designation by the Government of Japan as an anti-COVID-19 "ambassador"[17] for the Japanese people. Yet, Pepper's national identity is a bit murky and possibly takokuseki. The official homepage for Pepper does not specify its origin but implies a Japanese one by virtue of being on SoftBank's website[18]. The *IEEE Robots* database lists Pepper as a Japan-made robot[19]. Wikipedia lists both France and Japan[20]. The backstory of Pepper's development as reported in *IEEE Spectrum Magazine* suggests that its engineering was largely a French effort[21]. Still, a development report by its designers describes how Pepper was purposefully designed to embody visual cues linked to Japanese culture, such as large "kawaii" (cute) eyes [40]. As such, Pepper may be considered a multinational, i.e., takokuseki, or nationally ambiguous robot. Still, given its international fame and framing as a "Japanese" robot, we may expect people to attribute a Japanese nationality to it. This complexity in where Pepper was made, by whom and for whom, and how it has been framed in terms of its national identity presents an intriguing case study and point for comparison against the other mononational and less-known robots. We thus used Pepper to evaluate priming effects and perceptions of made-in and mukokuseki against actual knowledge (refer to 3.4.4). Even so, given this complexity, we decided to separate our analyses of Pepper from the other robots.

---

[4] https://www.deepl.com
[5] Due to copyright restrictions, we provide the stimuli figure on OSF: https://osf.io/tqvhz
[6] https://robots.ieee.org
[7] https://robots.ieee.org/robots/cody
[8] https://robots.ieee.org/robots/charli
[9] https://robots.ieee.org/robots/bandit
[10] https://robots.ieee.org/robots/octavia
[11] https://robots.ieee.org/robots/diegosan
[12] https://kawasakirobotics.com/eu-africa/products-robots/duaro1
[13] https://global.toyota/jp/download/34530903
[14] https://robots.ieee.org/robots/hrp4c
[15] https://robots.ieee.org/robots/pepper

Most images were sourced from the IEEE Robots database[22], except for duAro and T-HR3, which were sourced from the creators directly. As in previous research [51, 54], we cropped all images to the profile of the robot for consistency.

### 3.3.3 *Manipulation Check.*

We conducted a manipulation check to confirm the level of anthropomorphism for each robot in both phases, bundled together with the main questionnaire. We used the measure described in 3.4.1, analyzed as described in 3.5. Most robots were ascribed the expected Duffy category. However, the US iconic humanoid robot, Bandit, received lower anthropomorphism scores than expected. We replaced this robot with Octavia in the second phase. Nevertheless, Octavia received similar anthropomorphism scores, leading us to accept uneven mean differences between the Duffy categorizations. We report on the manipulation check results in full in 4.1.

---

[16] https://www.pepperparlor.com/en/
[17] https://www.softbank.jp/en/robot/
[18] https://www.softbank.jp/en/robot/#action_pepper
[19] https://robotsguide.com/robots/pepper
[20] https://en.wikipedia.org/wiki/Pepper_(robot)
[21] https://spectrum.ieee.org/how-aldebaran-robotics-built-its-friendly-humanoid-robot-pepper
[22] https://robots.ieee.org

## 3.4  MEASURES

### 3.4.1  *Perceived Anthropomorphism.*

We used the 5-point semantic differential anthropomorphism subscale from the Godspeed instrument [6]. The Godspeed is widely used instrument that has been translated into many languages, including Japanese. We used the anthropomorphism subscale to check our Duffy scale categorizations (Cronbach's alpha = .92).

### 3.4.2  *Perceived Nationality.*

We created a custom two-part instrument, in the absence of validated instruments. The first comprised one item to directly assess perceived nationality: "Nationality is a complex concept related to status as a member of a nation and geographic origin, as well as race, ethnicity, and culture. Do you think this robot has a nationality?" Respondents used a 3-point scale to indicate agreement and confidence (Yes, No, Unsure). We then asked participants to indicate what nationality, if any, the robot had, allowing for multiple selections from an expansive set of options to avoid bias and check the perceived multinationalism of Pepper. These included: North American, including Canada and the USA; Latin American; European; Middle Eastern; African or Caribbean; East Asian, including China and Japan; South Asian; a "robot" nationality, to account for liminal and ambiguous perceptions of robots as human enough for some national identity but not to the extent of having a *human* national identity; no nationality; and another nationality (please write).

### 3.4.3  *Perceived Similarity in Nationality.*

We created a custom two-part instrument reflecting direct and indirect measures of perceived national similarity and ingroup identification. We captured and triangulated direct and indirect measures as a means of rigour and quality assurance [37]. The direct measure used a 5-point Likert scale and asked: "Do you think this robot has the same nationality as you?" The indirect measure was placed within the Godspeed items as "Not Similar to Me" and "Similar to Me" poles.

### 3.4.4  Made-In Knowledge.

We created a custom qualitative instrument to check if respondents knew which countries the robots were made in, i.e., to account for priming effects, especially for Pepper. At the end of the survey, respondents were asked to write the country or nation for each robot, if they knew. They were able to skip each robot or the entire question.

### 3.4.5  Demographics.

We collected gender, age, and nationality at the end of the survey, to avoid potential priming effects [25] or stereotype threats [57]. Our response options were chosen based on recent guidelines for participant research, e.g., [30, 55], as well as the defaults of the SurveyMonkey platform. Gender options included woman/feminine, man/masculine, non-binary, gender fluid, trans*, another gender (please write), and prefer not to say. Age for American respondents was collected in 5-year increments from 18 to 75+. For Japanese respondents, age was collected by decade, e.g., 20s, 30s. Nationality options included USA, Japan, dual citizenship, and other (please write).

## 3.5  DATA ANALYSIS

Descriptive statistics were generated, including mean (M), standard deviation (SD), median (MD), interquartile range (IQR), and counts with percentages where appropriate. Perceived anthropomorphism items were averaged together to create one score. Inferential statistics were conducted for each robot, each national group of robots, and each national cohort, i.e., Japan and USA, according to the measures and hypotheses. Nonparametric statistics were used when normality and other checks failed. Made-in nationality responses were categorized and counted, and a grand score for the cohort was generated.

## 4  Results

We now present our results, organized by RQ and hypothesis alongside manipulation checks and other confirmative analyses. Descriptive statistics are found in Table 2 and Table 3.

*Table 2: Descriptive statistics for anthropomorphism by study phase, robot origin, and Duffy category.*

| Robot Origin | Duffy Category | Phase | Perceived Anthropomorphism | | | | Perceived Nationality | |
|---|---|---|---|---|---|---|---|---|
| | | | M | SD | MD | IQR | Matched | "Robot" |
| US | Industrial | 1 | 1.3 | 0.4 | 1 | 0.5 | 8 (7%) | 15 (14%) |
| US | Mechanical Humanoid | 1 | 1.8 | 0.8 | 1.5 | 1.2 | 7 (6%) | 18 (17%) |
| US | Iconic Humanoid | 1[a] | 2.2 | 0.9 | 2 | 1.2 | 21 (19%) | 19 (17%) |
| US | Humanoid | 1[b] | 3.4 | 0.9 | 3.5 | 1.2 | 69 (63%) | 6 (5%) |
| US | Industrial | 2 | 1.4 | 0.6 | 1.2 | 0.5 | 4 (4%) | 18 (19%) |
| US | Mechanical Humanoid | 2 | 1.8 | 0.8 | 1.5 | 1.2 | 5 (5%) | 25 (27%) |
| US | Iconic Humanoid | 2 | 2 | 0.7 | 2 | 1 | 5 (5%) | 28 (30%) |
| US | Humanoid | 2[b] | 3 | 0.9 | 3.2 | 1.5 | 44 (47%) | 3 (3%) |
| JP | Industrial | 1 | 1.3 | 0.5 | 1 | 0.2 | 7 (6%) | 15 (14%) |
| JP | Mechanical Humanoid | 1 | 1.7 | 0.7 | 1.8 | 1.2 | 7 (6%) | 19 (17%) |
| JP-FR | Iconic Humanoid | 1 | 2.4 | 0.8 | 2.5 | 1.2 | 23 (21%) | 20 (18%) |
| JP | Humanoid | 1[b] | 3.6 | 1 | 3.8 | 1.2 | 88 (81%) | 4 (4%) |
| JP | Industrial | 2 | 1.3 | 0.5 | 1 | 0.2 | 1 | 19 (20%) |
| JP | Mechanical Humanoid | 2 | 1.8 | 0.7 | 1.8 | 1.2 | 5 (5%) | 21 (23%) |
| JP-FR | Iconic Humanoid | 2 | 2.2 | 0.8 | 2 | 1.2 | 6 (6%) | 24 (26%) |
| JP | Humanoid | 2[b] | 3.3 | 0.9 | 3.5 | 1.2 | 73 (78%) | 7 (8%) |

*[a] Sig. diff. by respondent country, p < .05. [b] Sig. diff. by phase, p < .01 and p < .05.*

Table 3: Summarized descriptive statistics for key variables presented by robot origin.

| Robot Origin | Duffy Category | Perceived Anthropomorphism | | | | Perceived Nationality | | Perceived Similarity: Direct | | | | Perceived Similarity: Indirect | | | |
|---|---|---|---|---|---|---|---|---|---|---|---|---|---|---|---|
| | | M | SD | MD | IQR | Matched | "Robot" | M | SD | MD | IQR | M | SD | MD | IQR |
| US | Industrial | 1.4 | 0.5 | 1 | 0.5 | 12 (6%) | 33 (16%) | 1.1 | 1.1 | 1 | 2 | 1.2 | 0.6 | 1 | 0 |
| US | Mech. Hu. | 1.8 | 0.8 | 1.5 | 1.2 | 12 (6%) | 43 (21%) | 1.2 | 1.2 | 1 | 2 | 1.5 | 0.8 | 1 | 1 |
| US | Iconic Hu. | 2.1 | 0.8 | 2 | 1.2 | 26 (13%) | 47 (23%) | 1.4 | 1.3 | 1 | 2 | 1.7 | 0.8 | 1 | 1 |
| US | Humanoid | 3.2 | 0.9 | 3.4 | 1.5 | 113 (56%) | 9 (4%) | 2.1 | 1.5 | 2 | 2 | 2.6 | 1.2 | 3 | 2 |
| JP | Industrial | 1.3 | 0.5 | 1 | 0.2 | 8 (4%) | 34 (17%) | 1.0 | 1.1 | 1 | 2 | 1.1 | 0.6 | 1 | 0 |
| JP | Mech. Hu. | 1.8 | 0.7 | 1.8 | 1.2 | 12 (6%) | 40 (20%) | 1.3 | 1.2 | 1 | 2 | 1.5 | 0.8 | 1 | 1 |
| JP-FR | Iconic Hu. | 2.3 | 0.8 | 2.2 | 1.2 | 29 (14%) | 44 (22%) | 1.5 | 1.5 | 1 | 3 | 1.8 | 0.9 | 2 | 1 |
| JP | Humanoid | 3.4 | 0.9 | 3.8 | 1 | 161 (80%) | 11 (5%) | 2.0 | 1.5 | 2 | 2 | 3.0 | 1.2 | 3 | 2 |

*Mech.: Mechanical. Hu.: Humanoid.*

## 4.1   ANTHROPOMORPHISM MANIPULATION CHECK

We first considered whether the robots mapped onto the Duffy categories as expected, based on perceived anthropomorphism. In Phase 1, a t-test indicated a statistically significant difference between Japanese (M=2.0) and American (M=1.6) perceptions of Bandit, the US iconic robot, $t(106.6) = 2.6$, $p = .01$, 95% CI [.1, .7]. Due to this, we replaced Bandit with a different iconic robot made in the US but more like Pepper: Octavia. No statistically significant differences were found in Phase 2, confirming the fix.

We then conducted t-tests to compare Phases 1 and 2, treating Bandit and Octavia as the US iconic robot each time. We did not find a statistically significant difference, suggesting that, despite the Phase 1 results, participants across phases perceived Bandit and Octavia as roughly equal in terms of anthropomorphism. However, t-tests indicated a statistically significant difference for the US- and Japan-made humanoid robots, Diego-san and HRP-4C, between Phase 1 (M=3.4 and M=3.6) and Phase 2 (M=3.0 and M=3.3), $t(197) = 3.0$, $p = .003$, 95% CI [.1, .7] and $t(199.6) = 2.1$, $p = .04$, 95% CI [.02, .5], indicating lower anthropomorphism ratings in the second phase. Still, the means indicated high anthropomorphism overall. Moreover, Kruskal-Wallis tests and follow-up Dunn's tests indicated statistically significant differences among all US-made, $\chi2(3) = 338.0$, $p < .001$, $\eta p^2 = .42$, and Japan-made, $\chi2(3) = 413.4$, $p < .001$, $\eta p^2 = .52$, robots, indicating that all Duffy categories remained distinct for all robots across both national cohorts and phases. Notably, robots higher on the Duffy continuum were rated more human-like than those lower down (Figure 1).

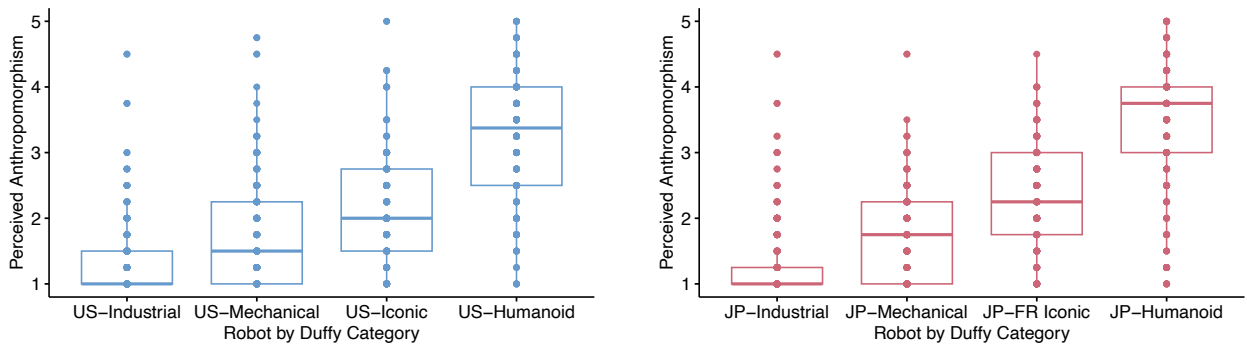

*Figure 1. Perceived anthropomorphism for the US (left) and JP plus JP-FR Pepper (right) robots.*

We thus confirmed that our categorization of the robots used in our subsequent analyses generally matched those of participants. We could combine the data collected in each phase.

## 4.2    NATIONALITY AND ANTHROPOMORPHISM (H1)

The distribution of attributions of nationality for the robots are presented in Table 4 and Figure 2. There were 1863 attributions.

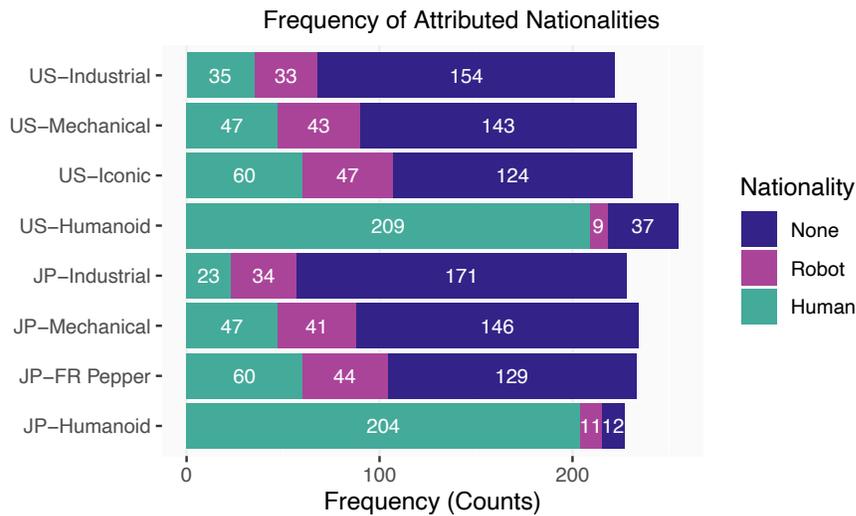

*Figure 2. Frequency of attributed nationalities.*

Participants tended to ascribe a nationality of some kind to the humanoid robots. Chi-squared tests indicated statistically significant differences in attributions of no nationality and some nationality ("robot" or human) to US-industrial, $\chi2(1) = 33.32$, $p < .001$, US-mechanical, $\chi2(1) = 12.16$, $p < .001$, US-humanoid, $\chi2(1) = 128.47$, $p < .001$, JP-industrial, $\chi2(1) = 57.00$, $p < .001$, JP-mechanical, $\chi2(1) = 14.38$, and JP-humanoid, $p < .001$, $\chi2(1) = 179.56$, $p < .001$. Descriptive statistics (Table 4) show fewer attributions of some nationality compared to no nationality for the industrial and mechanical robots, while the opposite was true for the humanoids. The iconic robots had roughly even attributions, with no statistically significant differences found for the US-iconic robot, $p = 26$, and JP-FR Pepper, $p = 10$.

*Table 4: Nationalities assigned to the robots by origin and mukokuseki, takokuseki, or made-in designation.*

| Rob. Ori. | Duffy Category | North Amer. | Latin Amer. | Europ. | Middle East. | Afric. | East Asian | Sum of Hu. Nat. | "Robot" | None | Desig. |
|---|---|---|---|---|---|---|---|---|---|---|---|
| US | Industrial | 16 (8%) | 5 (2%) | 7 (3%) | 2 (1%) | 1 | 7 (3%) | 35 | 33 | 154 | Made-in |
| US | Mech. Hu. | 23 (11%) | 2 (1%) | 17 (8%) | 2 (1%) | 0 | 8 (4%) | 47 | 43 | 143 | Made-in |
| US | Iconic Hu. | 18 (9%) | 7 (3%) | 8 (4%) | 10 (5%) | 8 (4%) | 16 (8%) | 60 | 47 | 124 | Tako. |
| US | Human. | 75 (37%) | 64 (32%) | 31 (15%) | 22 (11%) | 7 (3%) | 32 (16%) | 209 | 9 | 37 | Tako. |
| JP | Industrial | 9 (4%) | 2 (1%) | 4 (2%) | 3 (1%) | 2 (1%) | 5 (2%) | 23 | 34 | 171 | Muko. |
| JP | Mech. Hu. | 19 (9%) | 1 | 11 (5%) | 2 (1%) | 1 | 11 (5%) | 47 | 40 | 146 | Tako. |
| JP-FR | Iconic Hu. | 15 (7%) | 1 | 9 (4%) | 1 | 0 | 34 (17%) | 60 | 44 | 129 | Made-in |
| JP | Human. | 4 (2%) | 1 | 4 (2%) | 4 (2%) | 0 | 184 (91%) | 204 | 11 | 12 | Made-in |

*Note that respondents could select multiple options. Rob. Ori.: Robot Origin. Mech.: Mechanical. Hu.: Humanoid. Human.: Humanoid. Amer.: American. Europ.: European. East.: Eastern. Afric.: African. Hu.: Human. Nat.: Nationalities. Desig. Designation. Tako.: Takokuseki. Muko.: Mukokuseki.*

Participants also tended to ascribe a human, rather than "robot," nationality to the humanoid robots. Chi-squared tests indicated statistically significant differences in attributions of "robot" and human nationality to the US-humanoid, $\chi2(1)$ = 183.49, $p$ < .001, and JP-humanoid, $\chi2(1)$ = 173.25, $p$ < .001. Descriptive statistics (Table 4) show far more attributions of a human nationality for the humanoid robots, while the other types received roughly even numbers of "robot" and human nationality attributions.

Perceived level of anthropomorphism also related to ascriptions of nationality. Positive, statistically significant Kendall's tau correlations were found between perceived anthropomorphism and attribution of nationality for US-iconic, $\tau b$ = .18, $p$ = .003, US-humanoid, $\tau b$ = .24, $p$ < .001, JP-iconic, $\tau b$ = .22, $p$ < .001, and JP-humanoid, $\tau b$ = .19, $p$ = .001. Kruskal-Wallis tests indicated statistically significant differences for perceived anthropomorphism by attribution of nationality for US-iconic, $\chi2(2)$ = 9.68, $p$ = .008, $\eta p^2$ = .11, US-humanoid, $\chi2(2)$ = 17.78, $p$ < .001, $\eta p^2$ = .13, JP-industrial, $\chi2(2)$ = 8.34 $p$ = .016, $\eta p^2$ = .14, and JP-humanoid, $\chi2(2)$ = 10.23, $p$ = .006, $\eta p^2$ = .08. The results for the Japan-industrial robot appear to be an artifact of the small group size ("Yes" n=8 and anth. M=1.7 v. "No" n=181 and anth. M=1.2).

We now consider the liminal case of JP-FR iconic humanoid Pepper. A McNemar test did not find a statistically significant difference between ascriptions of human (n=60) and "robot" (n=44) nationalities, $\chi2(1)$ = 3.7, $p$ = .054. However, ones were found between human and none (n=129), $\chi2(1)$ = 21.89, $p$ < .001, and "robot" and none, $\chi2(1)$ = 41.76, $p$ < .001. This indicates relatively similar levels of among the two kinds of nationality ascriptions but significantly more ascriptions of no nationality to Pepper. For perceived anthropomorphism (M=2.3, SD=0.8, MD=2.2, IQR=1.2), a positive, statistically significant Kendall's tau correlation was found with ascriptions of nationality, $r_s(200)$ = .84, $p$ < .001, indicating that those who perceived Pepper as more anthropomorphic were also more inclined to ascribe it a nationality. A Kruskal-Wallis test indicated a statistically significant difference for perceived anthropomorphism by attribution of nationality, $\chi2(2)$ = 18.9, $p$ < .001, $\eta p^2$ = .13. This reflects the H1 results for the other robots.

The results suggest that the more humanlike, the more likely the robot was perceived as having a nationality, i.e., ***a made-in effect by anthropomorphism***, so we can ***accept hypothesis H1***. The number of ascriptions of some nationality rose with degree of anthropomorphism along the Duffy continuum. Ascriptions of a "robot" nationality were significantly low for the humanoids, while ascriptions of a human nationality were significantly high. The iconic robots represent a liminal case, with equal attributions of robot and human nationalities. In general, individuals who perceived higher degrees anthropomorphism in a given robot tended to also assign it a nationality of some kind, and humanoid robots tended to receive a human nationality overall.

## 4.3   MADE-IN KNOWLEDGE

Knowledge of where each robot was made could have influenced results. So, we asked participants to tell us if they knew which nation each robot was "made in." As the incorrect responses rates (Table 5) indicate, many made guesses, despite the instructions. Aside from the humanoid robots, most people did not know or did not guess correctly. A Chi-squared test did not find a statistically significant difference by country for number of guesses for US-made and JP-made robots, $p = .45$. This suggests that the Japanese and American respondents guessed in roughly equal numbers.

*Table 5: Nationalities assigned to the robots.*

| Robot Nation. | Duffy Category | US Correct | US Inc. | JP Correct | JP Inc. |
| --- | --- | --- | --- | --- | --- |
| US | Industrial | 2 | 2 | 4 | 10 |
| US | Mech. Hu. | 2 | 3 | 1 | 13 |
| US | Iconic Hu. | 3 | 7 | 3 | 13 |
| US | Humanoid | 20 | 16 | 6 | 11 |
| JP | Industrial | 0 | 2 | 2 | 14 |
| JP | Mech. Hu. | 1 | 2 | 2 | 13 |
| JP | Humanoid | 41 | 27 | 26 | 10 |

*Nation.: Nationality. Inc.: Incorrect. Mech.: Mechanical. Hu.: Humanoid.*

Chi-squared tests indicated significant statistical differences in the number of guesses by Duffy level by the US respondents for the US-made robots, $\chi2(3) = 34.78$, $p < .001$, but not the Japan-made robots, $p = .29$, as well as by the Japanese respondents for the US-made robots, $\chi2(2) = 74.42$, $p < .001$, and the Japan-made robots, $\chi2(3) = 38.4$, $p < .001$. The descriptive statistics (Table 5) show that this indicates a high tendency for guesses about humanoid robots, except by US respondents for the Japanese humanoid robot.

Let us now consider the multinational case of Pepper. A Chi-squared test indicated a statistically significant difference between US and Japanese respondent assignations to Pepper, $\chi2(1)$ = 21.42, $p$ < .001. Descriptive statistics (Table 6) suggest that Japanese respondents made more assignations than those from the US. Only twelve of 35 assignations (21.8%), all from the Japanese respondents, correctly stated Pepper's origin. 35 assignations (63.6%) by Japanese respondents indicated a Japan-only origin; seven of 16 (43.8%) assignations by US respondents did so, as well. This reflects **Pepper's association with Japan locally and abroad**. Yet, respondents made more incorrect guesses in general; moreover, most did not know or guess, leaving it blank.

*Table 6: Nationalities assigned to Pepper.*

| Nation | Japan | France | Both/SoftBank | Any |
|--------|-------|--------|---------------|-----|
| US | 7 | 0 | 0 | 16 |
| JP | 35 | 1 | 12 | 55 |

We will need to consider the results about Pepper's Japanese origin bias carefully, especially for the Japanese cohort. Still, the results support the relationship between high anthropomorphism and perceptions of nationality.

## 4.4   MADE-IN EFFECTS (H2)

A series of Chi-squared tests were conducted to compare each robot's made-in origins to participant ascriptions of nationality.

For the US-made robots, statistically significant differences were found when comparing nationalities (Table 4) for US-iconic, $\chi2(1)$ = 9.6, $p$ = .002, and US-humanoid, $\chi2(1)$ = 16.66, $p$ < .001, but not US-industrial, $p$ = .24, and US-mechanical, $p$ = .88. Participants ascribed the correct origin to the US-humanoid robot more frequently than no nationality, $\chi2(1)$ = 12.89, $p$ < .001. Moreover, those that attributed a human nationality to the industrial and mechanical robots tended to attribute the correct one: US-industrial, $\chi2(5)$ = 18, $p$ = .003, and US-mechanical, $\chi2(5)$ = 50.69, $p$ < .001. However, no difference was found among the human nationality attributions for the US-iconic robot, *p* = .083, and particularly for ascriptions of North American and East Asian nationalities, *p* = .73. This suggests a clear made-in effect for the US industrial and mechanical robots and a takokuseki or multinational effect for the iconic and humanoid robots.

For the Japan-made robots, a statistically significant difference favouring an East Asian ascription (Table 4) was found when comparing nationalities for the humanoid, $\chi2(1)$ = 131.84, *p* < .001. Participants also ascribed the correct origin to this robot more frequently than no nationality, $\chi2(1)$ = 150.94, *p* < .001. Unlike the US-made one, ascriptions of a human nationality varied for the JP-industrial robot, *p* = .13. Also, ascriptions of a human nationality for the JP-mechanical robot were spread between East Asian, North American, and European, with statistically significantly fewer counts for Latin American, Middle Eastern, and African, $\chi2(5)$ = 33.04, *p* < .001. This suggests a made-in affect for the JP-humanoid robot, a mukokuseki effect for the JP-industrial robot, and a takokuseki effect for the JP-mechanical robot.

As for Pepper, only three Japanese respondents correctly assigned it both European and East Asian nationalities. Still, attributions of human nationalities (Table 4) statistically significantly differed, $\chi2(5)$ = 84.86, *p* < .001. Notably, there were statistically significant differences favouring an East Asian nationality over a North American one, $\chi2(1)$ = 7.37, *p* = .007, and a European one, $\chi2(1)$ = 13.09, *p* < .001. This suggests a takokuseki effect, but not one that flatly affirms Pepper's FR-JP origins.

In summary, we can ***partially accept hypothesis H2***, that robots made in a certain nation will be more often assigned that nationality, i.e., a ***made-in effect by design****,* for the ***US-made non-humanoid robots and the Japan-made humanoid robot****.* The JP-industrial robot was mukokuseki. The other robots, including Pepper, appear to be takokuseki: multinational but not of any or every nation, indicating ambiguity or hybridity [32]. As mentioned in 4.3, we should consider the results for Pepper with some caution, given that it was known to nearly half of the Japanese participants and some of the American participants.

## 4.5    PERCEIVED SIMILARITY IN NATIONALITY (H3)

We then analyzed the relationship between the direct and indirect measures of perceived similarity and nationality.

For the US cohort, no statistically significant correlations were found for US-made robots or Japan-made robots.

For the Japanese cohort, statistically significant positive correlations were found between perceived as same nationality (MD=2, IQR=1.8) and perceived similarity (MD=1, IQR=1) for the US-made mechanical humanoid robot, τb = .25, *p* = .03, all of the Japan-made robots—the industrial robot, τb = .27, *p* = .03, MD=1, IQR=1 and MD=1, IQR=1, the mechanical humanoid, τb = .30, *p* < .001, MD=2, IQR=2 and MD=2, IQR=2, and the humanoid robot, τb = .31, *p* < .001, MD=4, IQR=1 and MD=4, IQR=1—and Pepper, τb = .24, *p* = .03, MD=3, IQR=2.8 and MD=3, IQR=2.8.

In sum, Japanese respondents tended to associate national similarity with self-similarity for Japanese-made robots and tended not with US-made ones. The US-made mechanical humanoid robot, CHARLI, may be a special case for the Japanese cohort, but we lack the qualitative accounts to explain it. No trends appeared to occur for American participants. This indicates a ***cultural effect*** on the relationship between perceiving a robot as having the same nationality as oneself (direct) and perceiving similarity between oneself and that robot (indirect) for Japan. However, as the medians show, this does not speak to the degree of similarity, only that participants used these measures similarly.

We now turn to degree of similarity. Figure 3 and Table 7 illustrate the results for participants ***who rated robots as having the same nationality as themselves***. Note that the counts vary widely by national cohort and robot, with some very low and others relatively high. We have used nonparametric statistics to account for this. Still, these results should be taken with caution.

For the US-made robots, a positive, statistically significant correlation was found by robot type and perceived similarity across cohorts, τb = .24, *p* < .001. A Kruskal-Wallis test confirmed differences in perceived similarity by type of robot across cohorts, $\chi2(3, n=65) = 12.29$, *p* = .007, $\eta p^2$ = .19. Dunn's tests indicated statistically significant differences between US-industrial and US-humanoid, *p* = .003 (*p* adjusted with the Benjamini-Hochberg method = .02). One between US-iconic and US-humanoid lost statistical significance after adjustment, *p* = .046, adj. *p* = .14. In sum, perceived similarity increased for the US robots along the Duffy continuum, especially between the lowest and highest.

We then divided by national cohort. A Kruskal-Wallis test found a statistically significant difference in perceived similarity by type of robot for US participants, $\chi2(3, n=42) = 10.33$, *p* = .016, $\eta p^2$ = .31, but not for Japanese participants, *p* = .45. For the US participants, Dunn's tests indicated no statistically significant differences, and a loss of statistically significance for US-industrial and US-humanoid, *p* = .044, adj. *p* = .13. This indicates that perceived similarity with US robots deemed to have the same nationality by the US cohort increased along the Duffy continuum, with results hampered by the small and varying group sizes.

For the Japan-made robots, a positive, statistically significant correlation was found by type of robot and perceived similarity across cohorts, τb = .33, *p* < .001. However, a Kruskal-Wallis test did not find a statistically significant difference in perceived similarity by type of robot across cohorts, *p* = .82, likely due to the low counts for JP-industrial (n=2) and JP-mechanical (n=4) compared to JP-humanoid (n=74). This hints at an increase in similarity for the Japan-made robots along the Duffy continuum.

As before, we also divided by national cohort for the Japan-made robots. Kruskal-Wallis tests did not find statistically significant differences in perceived similarity by type of robot for US participants, *p* = .14, or Japanese participants, *p* = .39. While descriptive statistics (Figure 3 and Table 7) suggest a similar trend to the US-made robots, the low and varying group sizes affected results. For example, only five Americans deemed the Japan-made robots to be American, and only one Japanese participant deemed JP-industrial as having the same nationality, compared to the JP-humanoid (n=70). Nevertheless, this again shows a connection between the high anthropomorphism and ascriptions of same nationality and similarity.

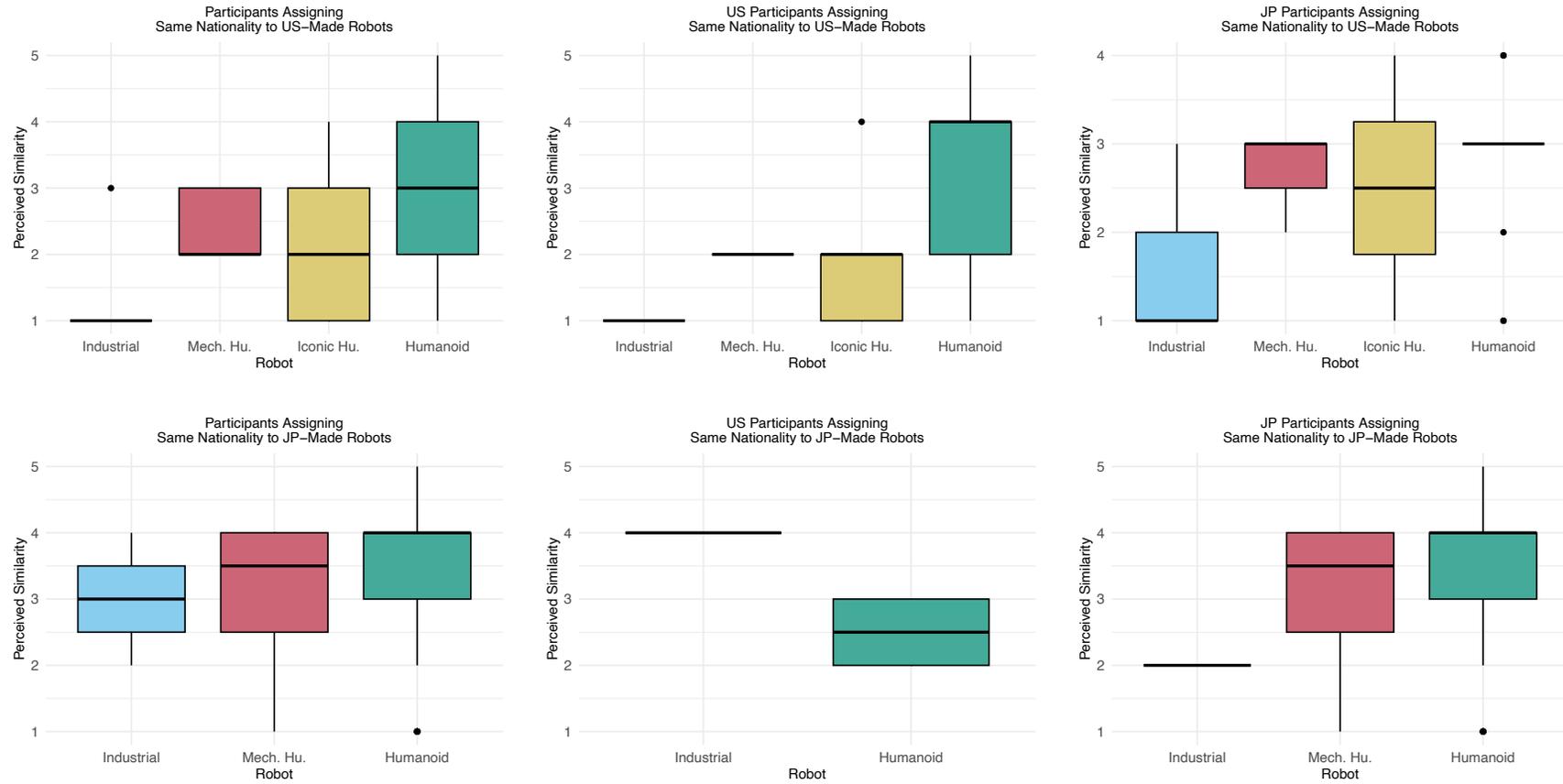

*Figure 3: Perceptions of similarity with US- and Japan-made robots by robot Duffy category and participant nationality. Mech. Hu.: Mechanical Humanoid. Iconic Hu.: Iconic Humanoid.*

*Table 7: Similarity-to-self when same nationality assigned.*

| Robot Nat. | Duffy Category | Particip. | | Perceived Similarity | | | |
|---|---|---|---|---|---|---|---|
| | | Nat. | n | M | SD | MD | IQR |
| US | Industrial | US | 2 | 1 | 0 | 1 | 0 |
| | | JP | 3 | 3 | 1.7 | 1 | 1 |
| US | Mech. Hu. | US | 2 | 2 | 0 | 2 | 0 |
| | | JP | 3 | 2.7 | 0.6 | 3 | 0.5 |
| US | Iconic Hu. | US | 5 | 2 | 1.2 | 2 | 1 |
| | | JP | 4 | 2.5 | 1.3 | 2.5 | 1.5 |
| US | Humanoid | US | 33 | 3.2 | 1.1 | 4 | 2 |
| | | JP | 13 | 2.8 | 0.9 | 3 | 0 |
| JP | Industrial | US | 1 | 4 | | 4 | |
| | | JP | 1 | 2 | | 2 | |
| JP | Mech. Hu. | US | 0 | | | | |
| | | JP | 4 | 3 | 1.4 | 3.5 | 1.5 |
| JP-FR | Iconic Hu. | US | 6 | 2.2 | 0.8 | 2 | 0.8 |
| | | JP | 24 | 2.4 | 1.2 | 2 | 3 |
| JP | Humanoid | US | 4 | 2.5 | 0.6 | 2.5 | 1 |
| | | JP | 70 | 3.4 | 1.2 | 4 | 1 |

*Nat.: Nationality. Particip.: Participant. Mech.: Mechanical. Hu.: Humanoid.*

As for Pepper, a positive, statistically significant correlation was found between agreement on same nationality and perceived similarity across cohorts, τb = .17, *p* = .004. A Kruskal-Wallis test did not find a statistically significant difference between the US (n=6, MD=2, IQR=.8) and Japanese (n=24, MD=2, IQR=3) respondents, *p* = .85 (Figure 4 and Table 7). This indicates that those who marked Pepper as having their own nationality had similar impressions of its similarity to themselves, regardless of nation, and that these were low in general.

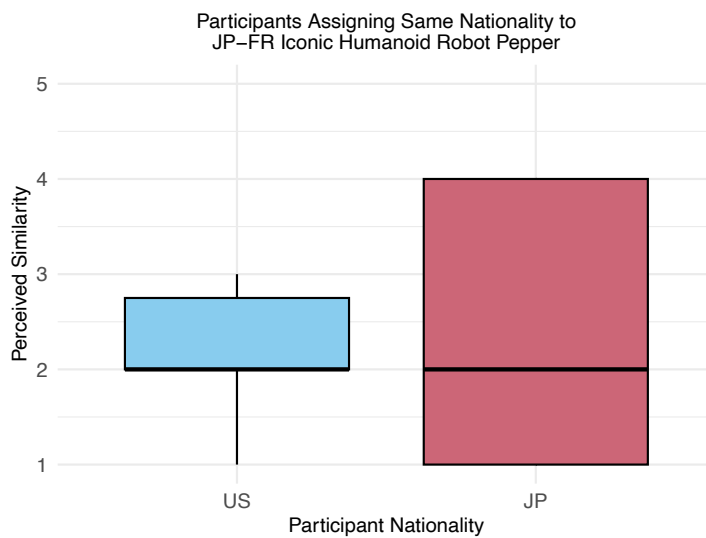

*Figure 4. Perceptions of similarity with Pepper, the JP-FR iconic humanoid robot, by participant nationality.*

Overall, perceived similarity with the robots marked by participants as having the same nationality as themselves increased with the anthropomorphism of the robot, despite the aforementioned cultural differences in how Japanese respondents used the direct and indirect measures. We can thus ***partially accept hypothesis H3***, that if people assigned a robot the same nationality as their own, then they perceived that robot as more like themselves, i.e., a ***made-in effect by similarity-to-self***, but ***only for robots of high anthropomorphism, i.e., humanoids***. We should interpret these results with caution, given the varying and small group sizes used in the statistics. Still, these results reflect the earlier findings on a link between anthropomorphism and perceptions of nationality. More pointedly, the cross-cultural results provide further nuance on the earlier results, supporting a takokuseki effect even for "made-in" robots. We discuss the possible explanations for this next, alongside the other results.

## 5   Discussion

The phenomenon of mukokuseki can occur for robots, but as our results indicate, this was generally not the case for most people. The standout exception was the Japan-made industrial robot. Anthropomorphism, perceptions of the robot's national identity and one's own nationality identity, and feelings of similarity interact in nuanced ways that point more often to perceptions of takokuseki, or multinationalism. Moreover, a "made-in" effect was found for robots along the Duffy continuum. We dig into these results in more depth and draw from previous research and theory to provide possible explanations and points for departure.

We discovered a clear mukokuseki effect only one robot, with few and equally varied ascriptions of nationality to the Japan-made industrial robot across national cohorts. We also found another variety of mukokuseki: takokuseki, or multiple but not comprehensive assignations of nationality for the US-made iconic robots (Bandit and Octavia), the US-made humanoid (Diego-san), the Japan-made mechanical humanoid robot, and the actually multinational iconic humanoid robot, Pepper, a collaboration between France and Japan. The US-made industrial and mechanical robots, as well as the Japan-made humanoid robot, HRP-4C, were subject to a "made-in" effect, somehow assigned the correct nationalities by most people, despite lack of actual knowledge by those people of each robot's origin. To answer RQ1, each effect occurred depending on the robot. To answer RQ2, anthropomorphism clearly played a large role but in unexpected ways. High anthropomorphism explained both the breakdown of a mukokuseki effect, i.e., the presence of a made-in effect, as for HRP-4C, but also the presence of mukokuseki and takokuseki effects, as for Diego-san. Prior knowledge may have influenced results in the case of Pepper but cannot account for the rest.

Diego-san was mukokuseki by way of takokuseki: ambiguity, plurality, or hybridity in national identity, receiving the range of human nationality ascriptions, but some more than others. Yet, the relative lack of African and Middle Eastern ascriptions is notable. This hints at a hidden association between national identity and race and/or ethnicity, given that African and Middle Eastern nations are generally associated with darker-skinned people. Specifically, these results allude to generalizations, if not stereotypes, about certain nationalities and race and/or ethnicity, especially in terms of skin colour, the primary anthropomorphic cue that separates Diego-san from the other types of robots along the Duffy scale. In other words, there appears to be a racial or ethnic component underlying these ascriptions. Whiteness is often presented as the default, including for products such as dolls [42]. Light skin is often positioned as ideal within the context of global white supremacy, a phenomenon known as colourism [11], which affects all people but perhaps especially non-white people and communities of colour, i.e., non-white American respondents and the Japanese respondents [29]. The apparent age of the robot, i.e., as a baby, may have also increased its ambiguity. In contrast, the Japan-made humanoid robot, HRP-4C, was designed to look like a teenaged Japanese girl. Future work can explore the intersections of age and race to tease out what national, ethic, racial, and/or cultural codes can be visually embedded into and perceptible within highly anthropomorphic robots.

We should also consider why the US- and Japan-made robots with the same relative anthropomorphism were not subject to the same effects, especially the "made-in" US-made industrial and mechanical humanoid robots. Perhaps these contain visual codes of US-ness that we cannot identity. The US is also a more racially and ethnic diverse nation than Japan, and we did not ask participants to self-report race or ethnicity. It is possible that the American respondents who identified with HRP-4C read racial/ethnic visual cues in the robot's appearance that matched their own, regardless of nationality. Certainly, participants did not know where these robots were made, as the freeform question results showed. Perhaps the global dominance of American industries and/or subtle anthropomorphic visual cues coded as white and/or American were perceived. More pointed qualitative work can confirm, refute, and/or expand on these possibilities.

The above leads us to raise a critical point about the limits and possible demerits of mukokuseki effects—pure mukokuseki or the multiplicitous takokuseki—for certain groups of people. An important limitation of our work is the lack of data on the racial and ethnic identities of our participants. Yet, recent statistics on the participant pools we used indicate demographic biases, with mTurk, for instance, being comprised of 79.9% white-identifying people and 9.1% Black-identifying people [34]. In the US context, at least, mukokuseki may simply reflect the status quo, leading naïve designers and researchers to create and employ robots that reflect those who have the greatest shares in representation and power and leaving uncharted others to the wayside. A crucial next step will be to capture participant demographics, especially those of a variety of racial and ethnic identities and confirm whether mukokuseki is truly "neutral" or "ambiguous," as well as when and to what extent takokuseki is "pluralistic."

We also found evidence of individual differences in self- and other-categorization, in line with previous work [36] and what we would expect for SIT, at least among people. Made-in robots might be mukokuseki for some, as when Japanese respondents ascribed an East Asian nationality to the US-made mechanical humanoid and iconic robots. Indeed, sampling at the national group level may occlude individual differences. Ambiguity, plurality, and hybridity in ascriptions of nationality, i.e., takokuseki, as found for the US-made Diego-san robot, may be better understood as an individual phenomenon rather than a Japanese or American group phenomenon. People may have different, even shifting orientations to facets of identity that then influence how social cues embedded in the visual design of robots, such as nationality, are perceived [4]. There are implications for diversity as Industry 5.0 approaches [2] and mass-produced robots that work with people enter our lives in ever greater numbers.

## 5.1   LIMITATIONS

We did not have access to whether mukokuseki was intended by the designers of the selected robots. As discussed, user perceptions may be the most critical in practice, which guided our methodological approach. Even so, future work should engage designers about the notion of mukokuseki in their work, such as through interviews, focus groups, or design jams. A clear next step for design and research practice would be exploring how to better align creator's expectations with user perceptions, in terms of national identity and other cues in the design of robots. A follow-up complementary trajectory would then be a matching study between expert intentions and user perceptions.

We were also unable to collect equivalent numbers of Japanese participants in the second phase due to unexpected limitations in the participant pool. Some groups were small because we could not select for certain responses in advance, e.g., shared national identity with a given robot. This particularly affected the results for H3. We hope that these results will guide the design of future work and sampling strategies. We also included a small selection of robots for only two nations and two national participant pools. Future work should explore a greater variety of robots and participant pools from other nations.

We could not control the number of data points for other variables, as well, notably whether and how many ascriptions of nationality would be made for each robot by each participant. This meant that some sample sizes were low and thus some statistical analyses were underpowered, e.g., the Chi-squared test for the human nationalities attributed to the JP-industrial robot (n = 24, power = .43). Since we could not account for this in advance, future work can learn from our data set and recruit more participants (double, in the case of this example).

While we conducted manipulation checks that, apart from the US iconic humanoid, matched our expectations, we acknowledge that our initial selection process could have been biased in subtle ways. Ultimately, the research team, with the notions of made-in, mukokuseki, and takokuseki in mind, decided on the baseline stimuli. The presence of such biases will be revealed in follow-up research with a broader range of robotic stimuli chosen by other researchers or as a result of a matching study with naïve subjects.

We did not explicitly ask about participant race and/or ethnicity, modeling our approach in line with how mukokuseki is defined in Japan. Future work should explore the nuances between race, ethnicity, nationality, and culture in robotic stimuli as well as in the responses from a diversity of participants across cultural contexts to ensure that mukokuseki and its kin, takokuseki, truly apply to everyone.

We did not collect qualitative data on the reasons behind participant ascriptions, which should be done in work.

## 6   Conclusion

People draw from human models to understand the robots that they see. This leads to mukokuseki, takokuseki, and made-in effects. Anthropomorphism is clearly at play. In the absence of familiarity with a robot, it is not clear what else explains a robot's relative mukokuseki-ness or made-in-ness. Future work will need to tease out possibilities, especially considering critical race theories and other work, within and outside of HRI, on how people perceive humanlikeness in robots.

## Acknowledgements

This work was funded by the department of the first author. We also thank Alexa Frank for introducing us to the concept of mukokuseki. We are truly grateful to the editors and reviewers for their service.